\def\year{2020}\relax
\documentclass{article} 
\usepackage{aaai20}  
\usepackage{times}  
\usepackage{helvet} 
\usepackage{courier}  
\usepackage[hyphens]{url}  
\usepackage{graphicx} 
\def\UrlFont{\rm}  
\usepackage{graphicx}  
\usepackage{floatrow}
\usepackage[algoruled]{algorithm2e}
\usepackage{graphicx}
\usepackage{amsmath}
\usepackage{hyperref}       
\usepackage{url}            
\usepackage{booktabs}       
\usepackage{amsfonts}       
\usepackage{nicefrac}       
\usepackage{microtype}      
\usepackage{mathtools}

\DeclareMathOperator*{\argmin}{arg\,min}

\DeclarePairedDelimiterX{\infdivx}[2]{(}{)}{%
  #1\;\delimsize\|\;#2%
}

\title{The Counterfactual $\chi$-GAN}

 \author{Amelia J. Averitt, \textsuperscript{\rm 1*} Natnicha Vanitchanant,\textsuperscript{\rm 1} Rajesh Ranganath,\textsuperscript{\rm 2} Adler J. Perotte\textsuperscript{\rm 1}  \\
 \textsuperscript{\rm 1}Biomedical Informatics, Columbia University, New York, NY \\
 \textsuperscript{\rm 2}Courant Institute, Center for Data Science, New York University, New York, NY \\ \textsuperscript{\rm *}\textit{Corresponding Author, aja2149@cumc.columbia.edu}
}

\begin{document}

\maketitle

\begin{abstract}
  Causal inference often relies on the counterfactual framework, which requires that treatment assignment is independent of the outcome, known as strong ignorability.  Approaches to enforcing strong ignorability in causal analyses of observational data include weighting and matching methods. Effect estimates, such as the \textit{average treatment effect} (ATE), are then estimated as expectations under the reweighted or matched distribution, $P$. The choice of $P$ is important and can impact the interpretation of the effect estimate and the variance of effect estimates. In this work, instead of specifying $P$, we learn a distribution that simultaneously maximizes coverage and minimizes variance of ATE estimates. In order to learn this distribution, this research proposes a generative adversarial network (GAN)-based model called the Counterfactual $\chi$-GAN (cGAN), which also learns feature-balancing weights and supports unbiased causal estimation in the absence of unobserved confounding. Our model minimizes the Pearson $\chi^2$-divergence, which we show simultaneously maximizes coverage and minimizes the variance of importance sampling estimates. To our knowledge, this is the first such application of the Pearson $\chi^2$-divergence. We demonstrate the effectiveness of cGAN in achieving feature balance relative to established weighting methods in simulation and with real-world medical data.
\end{abstract}

\section{Introduction}

Causal assessment often relies on the framework of \textit{counterfactual inference}. In this framework, each unit, $i$, has a \textit{potential outcome} given that they received a treatment and a potential outcome given that they received a control -- $Y_{1,i}$ and $Y_{0,i}$, respectively. This framework seeks to contrast the outcome, $Y$ for an individual under these two hypothetical states as shown in Eq. \ref{RubinModel} \cite{Rubin1974}.
\begin{equation}\label{RubinModel}
ITE = Y_1 - Y_0
\end{equation}
The effect of the treatment on the outcome can then summarized by calculating population-level effect estimates, such as the average treatment effect (ATE), which is defined as the expected difference in outcomes (Eq. \ref{ATE}).
\begin{equation} \label{ATE}
ATE =  \mathbb{E}[Y_{1}-Y_0] = \mathbb{E}[Y_{1}] - \mathbb{E}[Y_{0}] \\
\end{equation}
Estimating this requires access to the outcome for the state in which units were not assigned (i.e., $\mathbb{E}[Y_{0}|T=1]$ and $\mathbb{E}[Y_{1}|T=0]$). In practice, however, these true counterfactuals are never observed as a single population (or individual) cannot simultaneously be both treated and untreated. This is known as the 'fundamental problem of causal inference.' Therefore, approximations that employ more than one population are used as a proxy for these unobserved states \cite{holland1986}. These approximations seek to construct populations such that the observed ATE, $\hat{ATE}$, equals the true ATE that would arise from a counterfactual population. In other words, we seek an $\hat{ATE}$ that is \textit{unbiased}.
\begin{equation} \label{ATE_obs}
\hat{ATE} = \mathbb{E}[Y_{1}|T=1] - \mathbb{E}[Y_{0}|T=0] \\
\end{equation}
A decomposition of the ATE, demonstrates that a sufficient condition for unbiased $\hat{ATE}$ estimation is that $\mathbb{E}[Y_1|T=1)=E(Y_1|T=0)$ and $\mathbb{E}[Y_0|T=0)=E(Y_0|T=1)$ \cite{Kempthorne1955}. Within the counterfactual framework, this equality is central to the assumption of \textit{strong ignorability} (Eq. \ref{strong_ignorability}) \cite{Rosenbaum1983}. 
\begin{equation}\label{strong_ignorability}
Y_i(1), Y_i(0) \perp\!\!\!\perp T_i
\end{equation}
This assumption states a unit’s assignment to a treatment is independent of that unit’s potential outcomes, $Y_i$, and that treatment assignment is, therefore, ignorable. Causal claims borne from data that satisfy this requirement are regarded as unbiased as all confounding factors that could induce a dependence between $Y_i$ and $T_i$ are equally represented in the treatment and comparator arms \cite{Rubin1974}. Consequently, this means that the distribution of features is the same in both arms and features are said to be \textit{balanced}. Other assumptions, such as positivity and the Stable Unit Treatment Value Assumption (SUTVA), are also necessary and assumed to be true \cite{Rubin1980}.


Matching and weighting are popular pre-analysis manipulations to approximate the unconditional form of strong ignorability in observational populations. These methods create pseudo-populations in which the assumption is met without need for further manipulation \cite{Rubin1973}. This is opposed to methods of statistical adjustment, which occur peri-analysis, and approximate the conditional form of strong ignorability \cite{Leger1994}. Arguably, the most common strategy for weighting is the \textit{inverse probability of treatment weighting} (IPW) \cite{Thoemmes}, though other methods include the direct minimization of imbalance \cite{Gretton2009,Kallus2016,Kallus2017} or weighting by the odds of treatment, kernel weighting, and overlap weighting \cite{Rosenblatt1956,Hellerstein1999,Hazlett2016a,Li2016,Kallus2018}.

A commonality among these methods is that they implicitly or explicitly all specify a distribution function, $P$, that the expectation in Eq. \ref{ATE} is taken with respect to. This distribution is often the distribution associated with the treated ($p_1(x)$), the controls ($p_2(x)$), or a combination thereof (e.g. $\frac{1}{2}p_1(x) + \frac{1}{2}p_2(x)$). This choice of distribution can lead to high variance effect estimates in circumstances where there are regions of poor overlapping support between the treated and untreated populations. An effect of this is often observed in the context of IPW analyses with instability due to propensity scores near zero or one. \cite{Kang2007}. 

In this work, we instead construct an implicit distribution, $P$, that focuses on the regions of the sample space with significant overlap between the treated and untreated populations. Such a construction involves an inherent trade-off between coverage and variance. For example, mixture distributions that will be valid for a larger region of the sample space will also produce high variance estimates in the context of a fixed sample budget. In the context of infinite sample sizes and positivity, one could specify any distribution $P$ without concern for effect estimate variance. The mixture distribution of the treated and untreated populations would be a reasonable choice given a goal of maximizing coverage. However, in real-world settings with limited data, positivity may not be present and ATE estimates over such a distribution may be high variance in practice and theoretically invalid. In such a setting, valid estimates can only be made for subpopulations with significant distributional overlap. We formulate an approach that constructs a distribution $P$ for estimating Eq. \ref{ATE} that both maximizes coverage and minimizes variance. Informally, $P$ can be considered the distribution of a \emph{natural experiment} where the choice of treatment, $T$, is independent of potential confounders, $X$.

We propose the Counterfactual $\chi$-GAN (cGAN) that uses an adversarial approach to learn a distribution that trades off coverage and effect estimate variance for two or more observational study arms. This approach learns stable, feature balancing weights without reliance on the propensity score. The target distribution, $P$, is identified by minimizing the Pearson $\chi^2$-divergence between $P$ and the sampling distributions $Q_a$ for each study arm. To our knowledge, this is the first such application of the Pearson $\chi^2$-divergence. Because $P$ is being compared to all study arms, this encourages coverage, while, as we will show, the $\chi$-divergence inherently minimizes the variance of importance sampling estimates of the ATE.


This paper proceeds as follows: Section 2 defines the model and learning procedures, Section 3 presents an evaluation of this model through a simulation and an application to real-world clinical data, and finally, Section 4 discusses open issues, limitations of the model, and future work.

\section{The Model} \label{the_model}

We introduce the \emph{Counterfactual $\chi$-GAN} (cGAN), an adversarial approach to feature balance in causal inference that is based on importance sampling theory. Using an adversarial approach based on variational minimization based on the $f$-GAN, we minimize the sum of the Pearson $\chi^2$-divergences between a deep generative model and the sampling distributions from each arm of a study. We show that minimizing the $\chi^2$-divergence is equivalent, up to a constant factor, to minimizing the variance of importance sampling estimates to be made in approximating quantities such as ATEs. Similar to other weighting approaches, this approach assumes SUTVA, positivity, and no unmeasured confounders. In the following, $P$ is the constructed target distribution and $Q_a$ is the sampling distribution for each study arm.
 
\begin{figure}[ht]
\centering
  \includegraphics[width=1.0\textwidth]{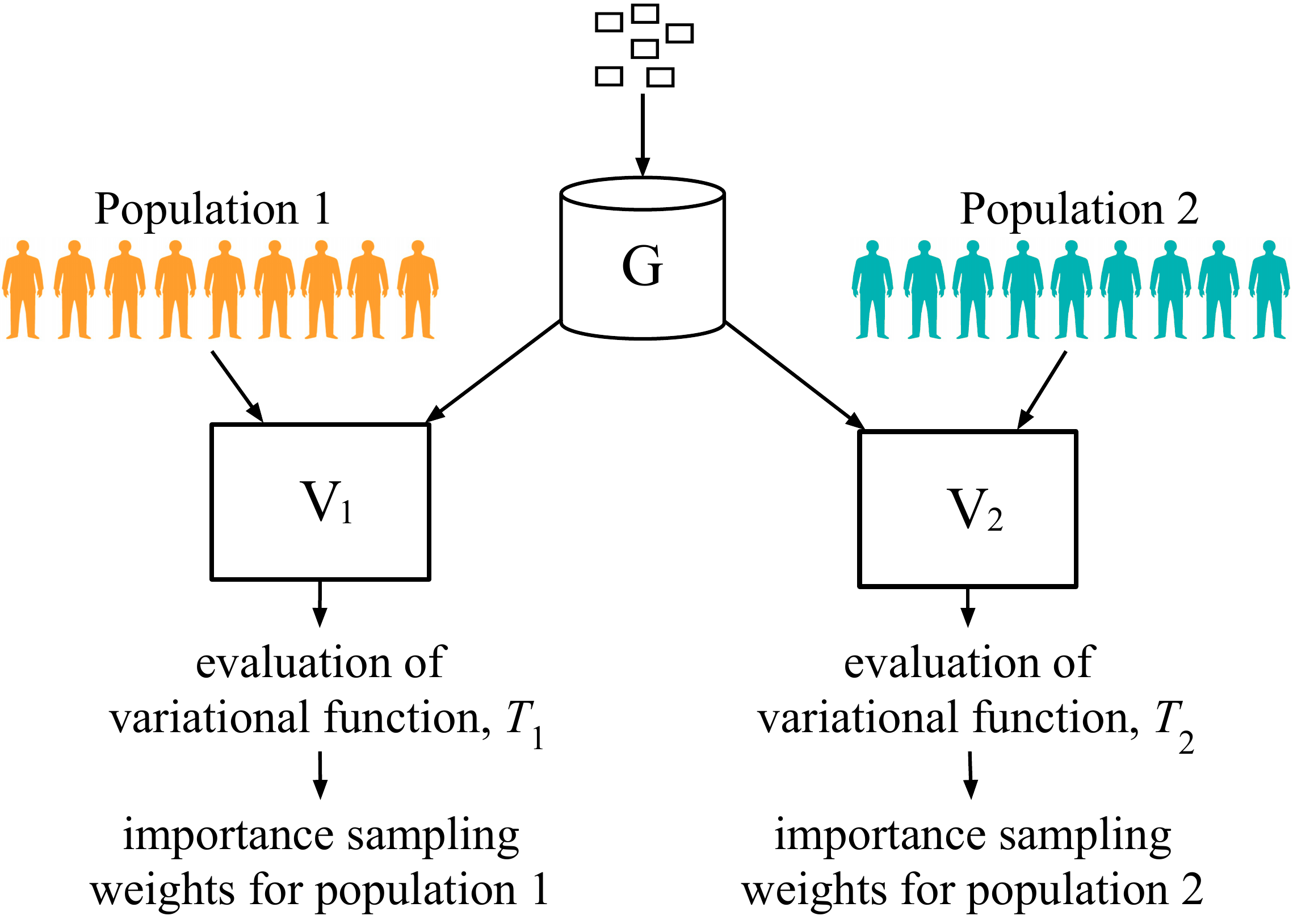}
  \caption{Architecture of Counterfactual $\chi$-GAN}
  \label{fig:cgan_schema}
\end{figure}
\vspace*{-5mm}
\paragraph{Importance Sampling and the $\chi^2$-divergence}
Importance sampling is a strategy for estimating expectations under an unknown target distribution given a known proposal distribution \cite{Hammersley1966}. Though the importance sampling has broader usage than our application, we focused on the use of importance sampling for estimation of the average treatment effect (ATE) because of its close relationship with the $\chi^2$ divergence. 
The importance sampling weight is defined as a likelihood ratio: the likelihood of an observation under the target distribution, $p(x)$ divided by the likelihood under the proposal distribution, $q(x)$. Weighted expectations based on the proposal distribution approximate unweighted expectations from the target distribution at shown in Eq. \ref{is_def}.
\begin{equation}
\mathbb{E}_q \left[ \frac{p(x)}{q(x)} \phi(x) \right] = \mathbb{E}_p \left[ \phi(x) \right]
\label{is_def}
\end{equation}
Consider the units in an arm of an observational study as being samples from such a proposal distribution. One strategy for obtaining unbiased expectations of treatment effects is to identify importance sampling weights for each arm that approximate expectations from a shared target distribution. However, this problem is underspecified given that we could choose any target distribution with the correct support.  In this work, we choose the target distribution that yields importance sampling approximations with smallest variance. Eq. \ref{is_var} shows the form for the variance of importance sampling estimates where $\phi(x)$ is the constant function. This choice is to make the formulation of the cGAN as outcome agnostic as possible. This form highlights its connection with the $\chi^2$-divergence, which has a function form as shown in Eq. \ref{chi_div}. This connection was previously noted in \cite{Dieng}. Therefore, the solution which minimizes the $\chi^2$-divergence would also minimize the variance expectations for unknown outcomes.  Of note, importance sampling is known to be a method that can produce high variance estimates, but since we will be minimizing the variance directly, this is less of a concern here.
\begin{equation}\label{is_var}
    \sigma^2_q = \frac{\mu^2}{n}\left( \int q(x)\left[\frac{p(x)^2}{q(x)^2} - 1\right]\right)dx 
  \end{equation} 
  \begin{equation}\label{chi_div}
    \chi^2 \infdivx{p}{q} = \int{q(x)\left[\frac{p(x)^2}{q(x)^2}-1\right]}dx
  \end{equation}
\paragraph{Likelihood Ratio ATE Estimation}
Typically, the expectation in the ATE is taken with respect to the original feature distribution, $q(x)$. Under cGAN-weighted data, expectations are taken with respect to the target distribution $p(x)$. As such, calculations of the ATE from the cGAN are not equivalent to what many would classically consider the ATE, but rather, is an ATE with respect to the new, learned feature distribution. We call this new estimate the $ATE_{p}$. This inequality is demonstrated in Equation \ref{ATE_pATE}. This set of equations shows that the typical ATE, $ATE_{q}$, is not equivalent to the expectation that we estimate, the $ATE_{p}$. 

\begin{equation}\label{ATE_pATE}
\begin{split}
ATE_{q} = &\mathbb{E}_{q(y_{1})}[y_{1}] - \mathbb{E}_{q(y_{0})}[y_{0}] \\
        = &\mathbb{E}_{q(x)}\mathbb{E}_{q(y_{1}|x)}[y_{1}|x] - \mathbb{E}_{q(x)}\mathbb{E}_{q(y_{0}|x)}[y_{0}|x] \\
        = &\mathbb{E}_{q(x)}\mathbb{E}_{q(y|x,t=1)}[y|x,t=1] - \\ & \mathbb{E}_{q(x)}\mathbb{E}_{q(y|x,t=0)}[y|x,t=0] \\
        = &\mathbb{E}_{q(x|t=1)}\frac{q(x)}{q(x|t=1)} \mathbb{E}_{q(y|x,t=1)}[y|x,t=1] - \\ & \mathbb{E}_{q(x|t=0)}\frac{q(x)}{q(x|t=0)} \mathbb{E}_{q(y|x,t=0)}[y|x,t=0]  \\
        \neq &\mathbb{E}_{q(x|t=1)}\frac{\textbf{p}(x)}{q(x|t=1)} \mathbb{E}_{q(y|x,t=1)}[y|x,t=1] - \\ 
        & \mathbb{E}_{q(x|t=0)}\frac{\textbf{p}(x)}{q(x|t=0)} \mathbb{E}_{q(y|x,t=0)}[y|x,t=0]
\end{split}
\end{equation}

Consider two distributions $Q_1$ and $Q_2$ that represent two arms of a study. It is possible to make unbiased $ATE_{p}$ estimates based on a single distribution, $P$, leveraging likelihood ratios/importance sampling weights as shown in Eq. \ref{ATE_LR}. 

\begin{equation} \label{ATE_LR}
ATE_{p} = \mathbb{E}_p[Y_{1}] - \mathbb{E}_p[Y_{0}] = \mathbb{E}_{q_1}\left[ \frac{p(x)}{q_1(x)}Y_{1}\right] - \mathbb{E}_{q_2}\left[ \frac{p(x)}{q_2(x)}Y_{0}\right]  \\
\end{equation}

We will leverage an approach based on adversarial learning to simultaneously maximizes coverage, minimizes the variance defined in Eq. \ref{is_var}, and directly estimates likelihood ratios, $\frac{p(x)}{q_1(x)}$ and $\frac{p(x)}{q_2(x)}$.


\paragraph{$f$-GAN}

The $f$-GAN framework provides a strategy for estimation and minimization of arbitrary $f$-divergences based on a variational divergence minimization approach \cite{Nowozin2016}. \begin{equation}\label{eq:f-bound}
\begin{split}
    D_f &\infdivx{P}{Q} = \int_{\mathcal{X}} q(x) \sup_{t\in dom_{f^*}} \left\{ t\frac{p(x)}{q(x)}-f^*(t) \right\} dx\\
    &\geq \sup_{T\in \mathcal{T}} \left( \int_{\mathcal{X}} p(x)T(x)dx -  \int_{\mathcal{X}} q(x) f^*(T(x))dx \right)\\
    &=\sup_{T\in\mathcal{T}} \left( \mathbb{E}_{x\sim P} [T(x)] - \mathbb{E}_{x\sim Q}[f^*(T(x))] \right)
    \end{split}
\end{equation}
where $T$ is a class of function such that $T: \mathcal{X} \rightarrow \mathbb{R}$, $f$ is the function that characterizes the $\chi^2$-divergence, $f(u)=(u-1)^2$, $f^*$ is the Fenchel conjugate of $f$, $f^*(t)=\frac{1}{4}t^2 + t$, and $P$ and $Q$ are probability distributions with continuous densities, $p(x)$ and $q(x)$. $T$ is typically a multi-layer neural network. This formulation lower bounds the $\chi^2$-divergence based on functions $T$, $P$, and $Q$ in such a way that unbiased noisy gradients of the lower bound can be easily obtained based on samples from $P$ and $Q$. In addition, the variational function, $T$, has a tight bound for $T^* = f^{\prime}\left(\frac{p(x)}{q(x)}\right)$ which is equivalent to $2\left(\frac{p(x)}{q(x)} - 1\right)$ in the case of the $\chi^2$-divergence. To respect the bounds of $T$ that result in valid likelihood ratios, we represent $T$ as a nonlinear transformation of an unbounded function $V$: $T(x) = g_f(V(x)) = -2 + log(1+e^{V(x)})$. The likelihood ratio, $\frac{p}{q}$, is easily derived from here and provides the importance sampling weights necessary for approximating expectations under $p(x)$ as shown in Eq. \ref{is_def}. 


\paragraph{The Counterfactual $\chi$-GAN}

The cGAN builds on importance sampling theory and extends the $f$-GAN framework to learn feature balancing weights through an adversarial training process. Previously, \cite{Tao2018} have explored importance weights from critics of divergence-based GAN models. However, unlike this method and other $f$-GANs where there is a generator, $G$ and a single variational function, the cGAN employs dual training from at least two variational functions (Figure \ref{fig:cgan_schema}).

Consider a set of $A$ treatments, each associated with one of $A$ populations, or arms of a study. Each population contains $N_a$ units and are drawn from an unknown and population-specific distribution $Q_a$. Based on the connection between the $\chi^2$-divergence and the variance of importance sampling estimates outlined above, our objective is to identify a target distribution that minimizes the $\chi^2$-divergence to all populations being compared: $\argmin_{p} \sum_{a=1}^{A} \chi^2 \left( p(x) \parallel q_a(x) \right) $.  This is the sum of the divergences between the generator and the unweighted treatment arms. It is minimized when $p(x)$ equals $q_a(x)$ for all $a$ and is directly proportional to the sum of the variances of importance sampling estimates under the target distribution, $P$, with proposals, $Q_a$. Because of the constant in Eq. \ref{is_var}, minimizing the $\chi^2$-divergence is equivalent to minimizing a normalized variance which weighs each population equally regardless of the number of units and the magnitude of the treatment effect, $\phi$.

\begin{algorithm}[ht]
\SetKwInOut{Input}{Input}
\SetKwInOut{Output}{Output}
 \Input{$(x_{1,1}$,...,$x_{1,N_1}$,...,$x_{A,N_A})$}
 \Output{$\theta$, $\omega_{1:A}$}
 
 Initialize $\theta$, $\omega_{1:A}$ and minibatch size, $M$.\\
 \While{$F(\theta, \omega_{1:A})$ not converged}{
    \For{$a\in (1,\ldots,A)$ treatment groups} {
        Sample a batch of noise samples, $z_{1:M}\sim p_g$, where $p_g$ is a prior distribution such as an isotropic Gaussian

        Sample minibatch of data, $x_{a,1:M}\sim q_{a}$
        
        Compute gradient w.r.t. variational function parameters
                \begin{center}
                $\nabla_{\omega_a} F = \sum_{m=1}^M \nabla_{\omega_a}  
                 (g_f(V_{\omega_a}(G_{\theta}(z_{m}))) - \frac{1}{4} g_f(V_{\omega_a}(x_{a,m}))^2 - g_f(V_{\omega_a}(x_{a,m})))$
                \end{center}
                
        Ascend the $\omega_a$ gradient according to a gradient-based optimizer
        
    } 
        
        
        Compute gradient w.r.t. generator parameters
            
        \begin{center}$\nabla_{\theta} F = \sum_{m=1}^M\sum_{a=1}^A \nabla_{\theta}\left[ g_f(V_{\omega_a}(G_{\theta}(z_{m})))\right]$\end{center}
            
        Descend the $\theta$ gradient according to a gradient-based optimizer
        
        Update $V_{\omega_a}$ and $G_{\theta}$ learning rates according to schedule
} 

\caption{Minibatch stochastic gradient cGAN optimization}
\label{alg:cGAN} 
\end{algorithm}

As a byproduct of minimizing this divergence, we will also identify a set of \emph{importance weights}, $w_{a,n}$, for each unit in each population that allows estimation of expectations from the same target distribution, $P$, thus satisfying the unconditional form of strong ignorability. Using these importance weights, expectations can be approximated as $\mathbb{E}_p[f] \approx \sum_{n=1}^{N_a} w_a \phi(x_{a,n})$ where $w_{a,n}=\frac{1}{c}\frac{p(x_{a,n})}{q_a(x_{a,n})}$, where $c=\sum_{n=1}^{N_a} \frac{p(x_n)}{q_a(x_n)}$ is an normalizing constant, $p$ is the density of the shared target distribution, $q_a$ is the density of the proposal distribution, and $x_{a,n}\sim Q_a$. Note that our strategy eliminates the need to explicitly evaluate $p\left(x_{a,n}\right)$ and $q_a(x_{a,n})$ as the likelihood ratio is estimated directly by the $f$-GAN. If desired, expectations can also be approximated using the sample-importance-resampling (SIR) algorithm where samples approximately distributed according to $p$ can be simulated by drawing samples from the weighted empirical distribution $\hat{q}_{a}(x)=\frac{1}{N_a}\sum_{n=1}^{N_a}w_{a,n}\delta(x-x_{a,n})$ \cite{Doucet2001}.

The objective function for the cGAN is shown in Eq. \ref{eq:F_loop} and is closely related to the objective defined in \cite{Nowozin2016}. $\theta$ parameterizes the generative model and $\omega_a$ parameterizes the variational model for each treatment arm, $a$. In our experiments, $V_{\omega_{a}}$ for all $a$ are neural networks that mirror discriminators in the traditional GAN framework and $P_\theta$ is a neural networks that mirrors the generator. Note that the generator in the original $f$-GAN framework is usually $Q_a$. In our case, to achieve the desired directionality of the $\chi^2$-divergence, the empirical distribution must be $Q_a$ and the generator must be $P$.  
\begin{multline}\label{eq:F_loop}
    F(\theta, \omega_{1:A})  = \sum_{t=1}^{A} \Bigg(  
    \mathbb{E}_{x \sim P_{\theta}} \big[ g_f(V_{\omega_{t}}(x)) \big] + \\
    \mathbb{E}_{x \sim Q_{a}} \big[ -\frac{1}{4} g_f(V_{\omega_a}(x))^2 - g_f(V_{\omega_a}(x)) \big] 
    \Bigg)
\end{multline}
Importance weights can be computed based on the fact that the bound in Eq. \ref{eq:f-bound} is tight for  $T^*(x)=f^{\prime}\left( \frac{p(x)}{q(x)} \right)$ where $f(u)=(u-1)^2$. We can therefore, approximate the desired importance weights as described in Eq. \ref{is_def} as $w_{a,n}=\frac{g_f(V_{\omega_a}(x_{a,n}))}{2}+1$ for all $a\in(1,\ldots,A)$ and $n\in(1,\ldots,N_a)$. Ultimately, the ATE can be estimated between any two treatment arms according to Eq. \ref{ATE_LR}. For example, the ATE between arms 1 and 2 could be estimated as $\hat{ATE} = \sum_{n=1}^{N_1}\left[w_{1,n}Y_{1,n}\right] - \sum_{n=1}^{N_2}\left[w_{2,n}Y_{2,n}\right]$.

\paragraph{Practical Considerations}
In the original GAN and $f$-GAN formulations the gradients for the generator is replaced with a related gradient that significantly speeds convergence of the model. Because our objective is minimization of the true $\chi^2$-divergence rather than perfect distributional matching, we do not employ this loss function trick but instead apply the gradient as derived from the loss function in Eq. \ref{eq:F_loop}.

Although it is the case that the domain of the Fenchel conjugate for the $\chi^2$-divergence is $\mathbb{R}$, we constrained it to $t \geq -2$ which produces valid likelihood ratios.


Gradient descent-based optimization of GANs is a notedly difficult task \cite{Mescheder2018,Arjovsky,Gulrajani}. Though many methods are proposed to stabilize training, we have found it sufficient to employ a set of algorithmic heuristics: (i) standardization of our data by the joint  mean and variance over all $A$ populations prior to training; (ii) periodically re-centering the distribution of each discriminator to a noisy estimate of the mean of the generator distribution. This re-centering is accomplished by setting the value of a vector that is added to the input of the discriminators. 


The approach for minibatch stochastic gradient descent for the cGAN is shown in Algorithm \ref{alg:cGAN}. The objective function \textit{F} (Eq. \ref{eq:F_loop}) is optimized by minimizing with respect to the parameters $\theta$ of the generator and maximizing with respect to the parameters $\omega_{1:A}$ of the discriminators. 

\paragraph{Related Work}

Causal inference with observational data has a rich literature that cuts across many disciplines \cite{Rosenbaum2002,Rubin1973,Rubin1974,Pearl2000} including machine learning \cite{Johansson2016,Kallus2018a,shalit2017estimating,ratkovic2014balancing,Schwab}. More specifically there have been several approaches to applying adversarial networks for counterfactual inference \cite{Kallus2018a,yoonJS18}. However, most existing methods for counterfactual inference are not directly comparable to the cGAN, as we aim to identify the most appropriate counterfactual distribution given the available data and maximize feature balance whereas most methods evaluate ATE estimation or ITE estimation directly.

In contrast to representational learning approaches and some GAN approaches, our approach does not rely on a predefined outcome to identify matched cohorts. The approach outlined in \cite{Kallus2018a} is the most similar in spirit to our approach but differs in that our objective directly minimizes the variance of expectations that might be used in ATE estimation, whereas \cite{Kallus2018a} minimizes a bound on the variance of the average treatment effect on the treated. As a result, there is no need for a regularizer, to perform cross-validation to select an appropriate level of regularization, or perform a constrained optimization over weights.

\section{Experiments}

To evaluate the cGAN, including its utility in practice, we present results of a simulation and applications to real-world medical data.

\subsection{Simulation} \label{simulation}

To evaluate the cGAN when the ground truth is known, we applied the model on simulated data of two populations/treatment arms, $A=2$. Each population was comprised of two subpopulations. Each subpopulation contained 10 features, drawn from a randomly generated multivariate normal distribution with a normal-Wishart prior distribution. Population 1 was composed of an equal number of samples (N=1000) from subpopulation A and subpopulation B; and Population 2 was composed of an equal number of samples from subpopulation A and subpopulation C (N=2000). By construction, subpopulation A is a latent population associated with a natural experiment, since it is part of both Population 1 and 2.  

\begin{figure*}[ht]
\centering
  \includegraphics[width=0.95\textwidth]{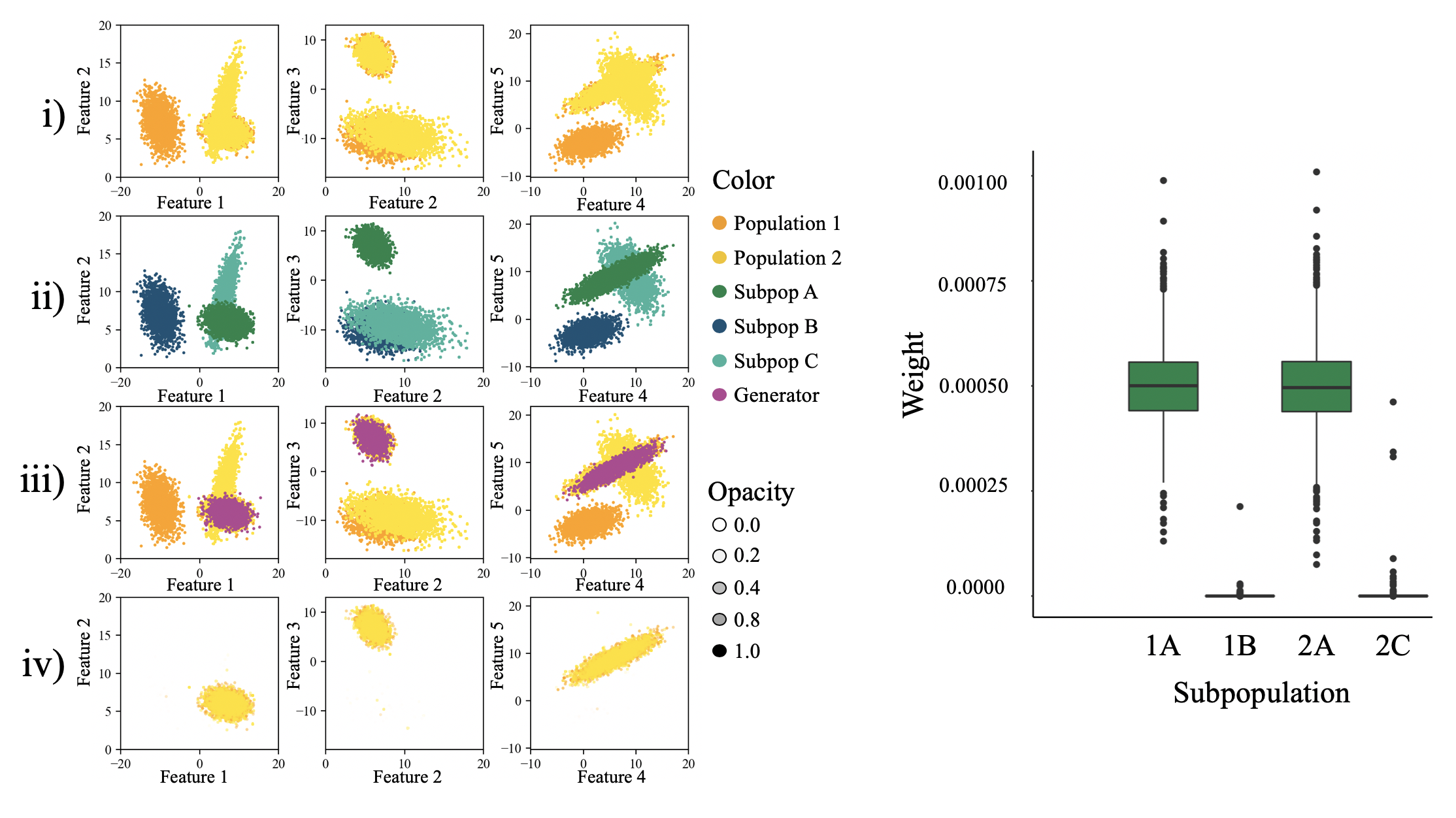}
  \caption{Simulation Results. \textit{Left:} Select features (i) by population of origin; (ii) with subpopulation A highlighted; (iii) samples from the generator; (iv) opacity adjusted by weight. \textit{Right:} Weights by subpopulation}
  \label{fig:simu_results}
\end{figure*}

Because our simulation deliberately constructs populations from a shared subpopulation distribution (A), we would expect points generated from this subpopulation to have higher weights. Intuitively, the variance of importance sampling estimates should be small for both treatment groups ($a=1$ and $a=2$) if the learned target distribution, $P_\theta$ is one that overlaps both populations maximally while excluding density unique to one group.

To better demonstrate how the cGAN supports counterfactual reasoning, we have additionally conducted an analysis of the average treatment effect (ATE) for our experiment with simulated data. We simulated a continuous outcome according to the subpopulation of origin -- Pop 1A $\sim$ Gaussian (60, 1); Pop 1B $\sim$ Gaussian (40, 1); Pop 2A $\sim$ Gaussian (-10, 1); Pop 2C $\sim$ Gaussian (10, 1). Under this outcome function, the estimate of average treatment effect (ATE) under the mixture distribution (of Pop 1 and Pop 2) is 50. When estimating the ATE under the overlapping subpopulation distribution -- those from Pop 1A and Pop 2A -- the ATE is 70. We applied weights from the cGAN and comparators to the simulated outcomes to assess the ability of the weighting methods to estimate one of the two ATEs. In addition, we also calculated the effective sample size (ESS), $n_{eff}$, using the Kish Method \cite{Kish1965}. The ESS may be used to determine the quality of a Monte Carlo approximations of importance sampling. The calculation of $n_{eff}$ can be found in the equation below, wherein $w$ are the weights.

\begin{equation*}
    n_{eff} = \frac{(\sum^{n}_{i=1}w_{i})^2}{\sum^{n}_{i=1}w^{2}_{i}}
    \label{neff}
\end{equation*}

To investigate (i) feature-balancing weights, (ii) the biasedness of ATE, and (iii) the ESS, a variety of comparator methods were implemented in addition to the cGAN . They include binary regression propensity score; generalized boosted modeling of propensity scores \cite{McCaffrey2004}; covariate-balancing propensity scores \cite{Imai2013}; non-parametric covariate-balancing propensity scores \cite{Fong2018}; entropy balancing weights \cite{Hainmueller2011}; empirical balancing calibration weights \cite{Chan}; optimization-based weights \cite{Zubizarreta2015}.


\paragraph{Results}
The results of our simulation is summarized in Figures \ref{fig:simu_results}. In the left hand-side of the Figure, the columns show the marginals of three pairs of continuous features. Row (i) shows the raw data, colored by which population units were drawn from. Row (ii) shows the same data as above, but coloring by subpopulation to highlight the overlapping distribution. Row (iii) shows a set of samples from the generator after training colored in blue. Row (iv) depicts the original data from Row (i) with the opacity of data points reflecting the importance weights. The right-hand side of the Figure shows the distribution of weights by subpopulation. Note that, in both Populations 1 and 2, the mean weights of units from subpopulation A have weights near $5x10^{-4}$, which is the uniform weight when 2000 units are in each population. Units from other subpopulations have near negligible weights, and would not meaningfully contribute to expectations in \ref{ATE_LR}.

\begin{table}[ht]
\centering
\begin{tabular}{l c}
Subpopulation & Mean Weight \\
\hline
1A          &   $4.997x10^{-4}$ \\ 
2B          &   $2.557x10^{-7}$ \\ 
2A          &   $4.992x10^{-4}$ \\
2C          &   $7.863x10^{-7}$ \\ 
\end{tabular}
\caption{Results of Application to Simulated Data. Mean cGAN-weight by subpopulation.}
\label{mean_weights}
\end{table}%

In the left-most figure, as you move down any column of feature pairs, it is apparent that points from the overlapping subpopulation A are both captured by the generator and assigned higher weights. This is confirmed by plotting the weights of data points by subpopulation (right-hand side of \ref{fig:simu_results}). Weights from subpopulations 1A and 2A are substantially higher than those from subpopulations 1B and 2C.

The results of this simulation further demonstrate that the ATE estimate from cGAN-weighted data is less biased than estimates from other weighting methods, given their respective targets. By construction, the causal effect of the comparable subpopulations is 70. cGAN-weighted data produced an ATE of 70.01. We see similarly good performance when inspecting the ESS. The cGAN has an ESS of 3870. Given that there are 4000 units that are comparable across the two arms (each subpopulation contains 2000 units), this is an appropriate estimate (Table \ref{ate_ess}).

\begin{table}[ht]
\begin{center}
\begin{tabular}{ l c c } 
 \hline
 \textbf{Weighting Method} & \textbf{ATE} & \textbf{ESS} \\ 
 \hline
 unweighted & 50.03 & 8000\\ 
 IPW & 92.00 & 6551 \\ 
 clipped IPW  & 87.24 & 6997 \\
 binary regression PS  & 92.00 & 6551 \\ 
 generalized boosted modeling PS & 84.51 & 7207 \\ 
 covariate balancing PS & 91.83 & 6686 \\ 
 non-parametric covariate balancing PS & 37.65 & 11 \\ 
 entropy balancing  & 104.13 & 65 \\ 
 empirical balancing calibration weights & 52.06 & 65 \\ 
 optimization-based weights & 52.07 & 114\\ 
 \textbf{cGAN} & \textbf{70.01}  & \textbf{3870} \\ 
 \hline
\end{tabular}
\end{center}
\caption{Results of Simulation. The average treatment effect and effective sample size (ESS) after application of weighting methods from the Counterfactual $\chi$-GAN and comparators.}
\label{ate_ess}
\end{table}

\subsection{Clinical Data} \label{experimentation}

We additionally applied the cGAN to an experiments using real-world clinical data from a large, academic medical center. For this experiment, we constructed the treatment and comparator cohorts according to the protocol and indication of a published randomized clinical trial. The experiment compares sitagliptin and glimepiride in elderly patients with Type II Diabetes Mellitus (N=144 per arm) \cite{Hartley2015}. We present the 37 most frequent clinical measurements from the electronic health record. 

We evaluate the ability of the cGAN to improve feature balance by comparing the Absolute Standardized Difference of Means (ASDM) between the treatment and comparator cohorts under different weighting methods. the ASDM is a popular method of assessing cohort similarity, with a lower metric corresponding to improved feature balance. The ASDM is presented for the cGAN and the comparator weighting methods mentioned in the simulation. Under the clipped-IPW procedure, propensity scores greater than 90th percentile and less than 10th percentile are assigned to the values of the percentiles at 90th and 10th, respectively \cite{Cole2008}.

\begin{figure*}[t]
\centering
  \includegraphics[width=1.0\textwidth]{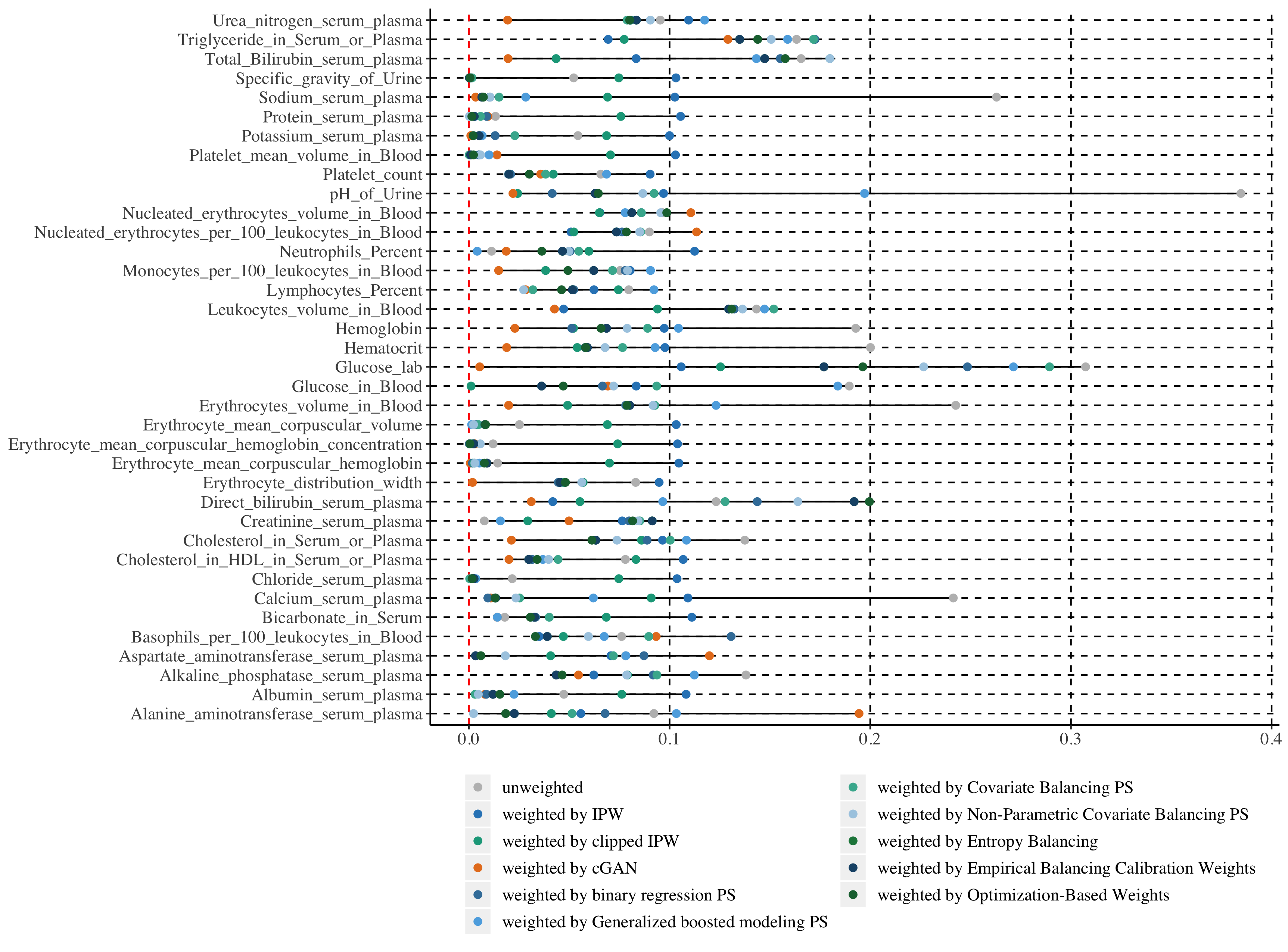}
  \vspace*{-5mm}
  \caption{Absolute standardized difference of the means (ASDM) of real-world clinical features after application weighting methods from the Counterfactual $\chi$-GAN and comparators.}
  \label{exp_asdm}
\end{figure*}

\paragraph{Results} The ASDM for the clinical cohorts is presented in Figure \ref{exp_asdm}. These findings are summarized by the mean ASDM over all features, under the varying weighting methods in Table \ref{ASDM_table_exp}. cGAN improved mean ASDM from the unweighted cohort and improved feature balance the most among all evaluated methods. Note that this task is particularly challenging due to the high dimensionality of the data and small study size. 

The results of this experiment can be found in Figure \ref{exp_asdm} and Table \ref{ASDM_table_exp}. They demonstrate that cGAN-weighting achieves better feature balance than comparator methods. 

\begin{table}[h]
\begin{center}
\begin{tabular}{ l c } 
 \hline
 \textbf{Weighting Method} & \textbf{ASDM} \\ 
 \hline
 unweighted & 0.1103 \\ 
 IPW & 0.0876 \\ 
 clipped IPW  & 0.0631 \\
 binary regression PS  & 0.0625 \\ 
 generalized boosted modeling PS & 0.0749 \\ 
 covariate balancing PS & 0.0681 \\ 
 non-parametric covariate balancing PS & 0.0596 \\ 
 entropy balancing  & 0.0524 \\ 
 empirical balancing calibration weights & 0.0524 \\ 
 optimization-based weights & 0.0536 \\ 
 \textbf{cGAN} & \textbf{0.0364} \\ 
 \hline
\end{tabular}
\end{center}
\caption{Results of Application to Clinical Data. Absolute standardized difference of the means (ASDM) of real-world clinical features after application weighting methods from the Counterfactual $\chi$-GAN and comparators.}
\label{ASDM_table_exp}
\end{table}

\vspace*{-5mm}

\section{Discussion}
In this paper, we introduce the Counterfactual $\chi$-GAN. It is a deep generative model for feature balance that minimizes the variance of importance sampling estimates of treatment effects. We leverage the $f$-GAN framework for estimating the $\chi^2$-divergence and likelihood ratios necessary for achieving this.

The experiments presented here suggest that cGAN is an effective method of learning feature balancing weights to support counterfactual inference. If we assume that all potentially confounding variables are observed, the superiority of cGAN in learning balancing weights, suggests that ATE borne from cGAN-weighted cohorts would be less biased than those estimates generated from traditional weighting methods.

The application of the model to real-world EHR data, demonstrates that this method could provide an alternative means to causal estimation from observational data when the assumptions of no unobserved confounding, positivity, and SUTVA are met. Our experiments suggest that the flexibility of our framework produces improved feature balance relevant for valid causal estimates. 
This method does, however, come with limitations. Training of the model is completed via backpropogation. As such, this method is only suitable for fully differentiable functions. Therefore, matching based on a combination of discrete and continuous values poses a challenge.  In addition, GANs are well known for their instability and lack of objective measures for convergence. This work shares those limitations. In future work, we will explore an extension of the cGAN which accommodates discrete data types and overcome the many current limitations of GANs.

\subsubsection*{Acknowledgments}

This research is supported by grants R01LM009886-10 and T15LM007079 from The National Library of Medicine.

\bibliography{main.bib}

\begin{thebibliography}{}

\bibitem[\protect\citeauthoryear{Arjovsky and Bottou}{2017}]{Arjovsky}
Arjovsky, M., and Bottou, L.
\newblock 2017.
\newblock Towards principled methods for training generative adversarial
  networks.

\bibitem[\protect\citeauthoryear{Chan, Yam, and Zhang}{2016}]{Chan}
Chan, K. C.~G.; Yam, S. C.~P.; and Zhang, Z.
\newblock 2016.
\newblock {Globally Efficient Nonparametric Inference of Average Treatment
  Effects by Empirical Balancing Calibration Weighting}.
\newblock Technical report, University of Washington.

\bibitem[\protect\citeauthoryear{Cole and Hern{\'{a}}n}{2008}]{Cole2008}
Cole, S.~R., and Hern{\'{a}}n, M.~A.
\newblock 2008.
\newblock {Constructing inverse probability weights for marginal structural
  models.}
\newblock {\em American Journal of Epidemiology} 168(6):656--64.

\bibitem[\protect\citeauthoryear{Dieng \bgroup et al\mbox.\egroup
  }{2017}]{Dieng}
Dieng, A.~B.; Tran, D.; Ranganath, R.; Paisley, J.; and Blei, D.~M.
\newblock 2017.
\newblock {Variational Inference via $\chi$ Upper Bound Minimization}.
\newblock In {\em NIPS}.

\bibitem[\protect\citeauthoryear{Doucet, Freitas, and
  Gordon}{2001}]{Doucet2001}
Doucet, A.; Freitas, N.; and Gordon, N.
\newblock 2001.
\newblock {An Introduction to Sequential Monte Carlo Methods}.
\newblock In {\em Sequential Monte Carlo Methods in Practice}. New York, NY:
  Springer New York.
\newblock  3--14.

\bibitem[\protect\citeauthoryear{Fong, Hazlett, and Imai}{2018}]{Fong2018}
Fong, C.; Hazlett, C.; and Imai, K.
\newblock 2018.
\newblock {Covariate balancing propensity score for a continuous treatment:
  Application to the efficacy of political advertisements}.
\newblock {\em The Annals of Applied Statistics} 12(1):156--177.

\bibitem[\protect\citeauthoryear{Gretton \bgroup et al\mbox.\egroup
  }{2009}]{Gretton2009}
Gretton, A.; Smola, A.; Huang, J.; Schmittfull, M.; Borgwardt, K.; Schölkopf,
  B.; Candela, J.; Sugiyama, M.; Schwaighofer, A.; and Lawrence, N.
\newblock 2009.
\newblock Covariate shift by kernel mean matching.
\newblock {\em Dataset Shift in Machine Learning, 131-160 (2009)}.

\bibitem[\protect\citeauthoryear{Gulrajani \bgroup et al\mbox.\egroup
  }{2017}]{Gulrajani}
Gulrajani, I.; Ahmed, F.; Arjovsky, M.; Dumoulin, V.; and Courville, A.
\newblock 2017.
\newblock {Improved Training of Wasserstein GANs}.
\newblock In {\em NIPS}.

\bibitem[\protect\citeauthoryear{Hainmueller}{2011}]{Hainmueller2011}
Hainmueller, J.
\newblock 2011.
\newblock {Entropy Balancing for Causal Effects: A Multivariate Reweighting
  Method to Produce Balanced Samples in Observational Studies}.
\newblock {\em Political Analysis} 16:25--46.

\bibitem[\protect\citeauthoryear{Hartley \bgroup et al\mbox.\egroup
  }{2015}]{Hartley2015}
Hartley, P.; Shentu, Y.; Betz-Schiff, P.; Golm, G.~T.; Sisk, C.~M.; Engel,
  S.~S.; and Shankar, R.~R.
\newblock 2015.
\newblock {Efficacy and Tolerability of Sitagliptin Compared with Glimepiride
  in Elderly Patients with Type 2 Diabetes Mellitus and Inadequate Glycemic
  Control: A Randomized, Double-Blind, Non-Inferiority Trial}.
\newblock {\em Drugs {\&} Aging} 32(6):469--476.

\bibitem[\protect\citeauthoryear{Hazlett}{2016}]{Hazlett2016a}
Hazlett, C.
\newblock 2016.
\newblock {Kernel Balancing: A Flexible Non-Parametric Weighting Procedure for
  Estimating Causal Effects}.
\newblock {\em SSRN}.

\bibitem[\protect\citeauthoryear{Hellerstein and
  Imbens}{1999}]{Hellerstein1999}
Hellerstein, J.~K., and Imbens, G.~W.
\newblock 1999.
\newblock {Imposing moment restrictions from auxiliary data by weighting}.
\newblock {\em Review of Economics and Statistics} 81:1--14.

\bibitem[\protect\citeauthoryear{Holland}{1986}]{holland1986}
Holland, P.~W.
\newblock 1986.
\newblock {Statistics and Causal Inference}.
\newblock {\em JASA} 81(396):945--960.

\bibitem[\protect\citeauthoryear{Imai and Ratkovic}{2013}]{Imai2013}
Imai, K., and Ratkovic, M.
\newblock 2013.
\newblock {Covariate balancing propensity score}.
\newblock Technical report, Harvard University.

\bibitem[\protect\citeauthoryear{Johansson, Shalit, and
  Sontag}{2016}]{Johansson2016}
Johansson, F.~D.; Shalit, U.; and Sontag, D.
\newblock 2016.
\newblock Learning representations for counterfactual inference.

\bibitem[\protect\citeauthoryear{Kallus}{2016}]{Kallus2016}
Kallus, N.
\newblock 2016.
\newblock {Causal inference by minimizing the dual norm of bias: Kernel
  matching {\&} weighting estimators for causal effects}.
\newblock {\em CEUR Workshop Proceedings} 1792:18--28.

\bibitem[\protect\citeauthoryear{Kallus}{2017}]{Kallus2017}
Kallus, N.
\newblock 2017.
\newblock {A Framework for Optimal Matching for Causal Inference}.
\newblock In {\em AISTATS},  372--381.

\bibitem[\protect\citeauthoryear{Kallus}{2018a}]{Kallus2018a}
Kallus, N.
\newblock 2018a.
\newblock Deepmatch: Balancing deep covariate representations for causal
  inference using adversarial training.

\bibitem[\protect\citeauthoryear{Kallus}{2018b}]{Kallus2018}
Kallus, N.
\newblock 2018b.
\newblock {Optimal a priori balance in the design of controlled experiments}.
\newblock {\em J. R. Statist. Soc. B} 80(1):85--112.

\bibitem[\protect\citeauthoryear{Kang and Schafer}{2007}]{Kang2007}
Kang, J. D.~Y., and Schafer, J.~L.
\newblock 2007.
\newblock {Demystifying Double Robustness: A Comparison of Alternative
  Strategies for Estimating a Population Mean from Incomplete Data}.
\newblock {\em Statistical Science} 22(4):523--539.

\bibitem[\protect\citeauthoryear{Keele and Zubizarreta}{2014}]{Zubizarreta2015}
Keele, L., and Zubizarreta, J.
\newblock 2014.
\newblock {Optimal Multilevel Matching in Clustered Observational Studies : A
  Case Study of the School Voucher System in Chile}.
\newblock {\em arXiv preprint arXiv:1409.8597}  1--37.

\bibitem[\protect\citeauthoryear{Kempthorne}{1955}]{Kempthorne1955}
Kempthorne, O.
\newblock 1955.
\newblock {The randomizaton theory of experimental inference}.
\newblock {\em Journal of American Statistics} 50:946--967.

\bibitem[\protect\citeauthoryear{Kish}{1965}]{Kish1965}
Kish, L.
\newblock 1965.
\newblock {Survey Sampling}.
\newblock In {\em Survey Sampling}. New York: Wiley.
\newblock chapter Chapter 14.

\bibitem[\protect\citeauthoryear{Leger}{1994}]{Leger1994}
Leger, A. S.~S.
\newblock 1994.
\newblock {\em {Statistical Models in Epidemiology}}, volume~48.
\newblock Oxford University Press.

\bibitem[\protect\citeauthoryear{Li, Morgan, and Zaslavsky}{2018}]{Li2016}
Li, F.; Morgan, K.~L.; and Zaslavsky, A.~M.
\newblock 2018.
\newblock {Balancing Covariates via Propensity Score Weighting}.
\newblock {\em Journal of the American Statistical Association}
  113(521):390--400.

\bibitem[\protect\citeauthoryear{McCaffrey, Ridgeway, and
  Morral}{2004}]{McCaffrey2004}
McCaffrey, D.~F.; Ridgeway, G.; and Morral, A.~R.
\newblock 2004.
\newblock {Propensity Score Estimation With Boosted Regression for Evaluating
  Causal Effects in Observational Studies.}
\newblock {\em Psychological Methods} 9(4):403--425.

\bibitem[\protect\citeauthoryear{Mescheder, Geiger, and
  Nowozin}{2018}]{Mescheder2018}
Mescheder, L.; Geiger, A.; and Nowozin, S.
\newblock 2018.
\newblock Which training methods for gans do actually converge?

\bibitem[\protect\citeauthoryear{Muller}{1966}]{Hammersley1966}
Muller, M.~E.
\newblock 1966.
\newblock {\em {Review: J. M. Hammersley, D. C. Handscomb, Monte Carlo Methods
  ; Yu. A. Shreider, Methods of Statistical Testing/Monte Carlo Method}},
  volume~37.
\newblock Springer Netherlands.

\bibitem[\protect\citeauthoryear{Nowozin, Cseke, and
  Tomioka}{2016}]{Nowozin2016}
Nowozin, S.; Cseke, B.; and Tomioka, R.
\newblock 2016.
\newblock {f-GAN: Training Generative Neural Samplers using Variational
  Divergence Minimization}.
\newblock In {\em NIPS},  271--279.

\bibitem[\protect\citeauthoryear{Pearl}{2000}]{Pearl2000}
Pearl, J.
\newblock 2000.
\newblock {\em {Causality}}.
\newblock Cambridge, England: Cambridge University Press.

\bibitem[\protect\citeauthoryear{Ratkovic}{2014}]{ratkovic2014balancing}
Ratkovic, M.
\newblock 2014.
\newblock Balancing within the margin: Causal effect estimation with support
  vector machines.
\newblock {\em Department of Politics, Princeton University, Princeton, NJ}.

\bibitem[\protect\citeauthoryear{Rosenbaum and Rubin}{1983}]{Rosenbaum1983}
Rosenbaum, P.~R., and Rubin, D.~B.
\newblock 1983.
\newblock {Assessing Sensitivity to an Unobserved Binary Covariate in an
  Observational Study with Binary Outcome}.

\bibitem[\protect\citeauthoryear{Rosenblatt}{1956}]{Rosenblatt1956}
Rosenblatt, M.
\newblock 1956.
\newblock {Remarks on Some Nonparametric Estimates of a Density Function}.
\newblock {\em The Annals of Mathematical Statistics} 27(3):832--837.

\bibitem[\protect\citeauthoryear{Rubin}{1973}]{Rubin1973}
Rubin, D.~B.
\newblock 1973.
\newblock {The Use of Matched Sampling and Regression Adjustment to Remove Bias
  in Observational Studies}.
\newblock {\em Biometrics} 29(1):185.

\bibitem[\protect\citeauthoryear{Rubin}{1974}]{Rubin1974}
Rubin, D.~B.
\newblock 1974.
\newblock {Estimating causal effects of treatments in randomized and
  nonrandomized studies.}
\newblock {\em Journal of Educational Psychology} 66(5):688--701.

\bibitem[\protect\citeauthoryear{Rubin}{1980}]{Rubin1980}
Rubin, D.
\newblock 1980.
\newblock {Randomization analysis of experimental data: The Fisher
  randomization test comment}.
\newblock {\em JASA} 75(371):591--593.

\bibitem[\protect\citeauthoryear{Schwab, Linhardt, and Karlen}{2018}]{Schwab}
Schwab, P.; Linhardt, L.; and Karlen, W.
\newblock 2018.
\newblock Perfect match: A simple method for learning representations for
  counterfactual inference with neural networks.

\bibitem[\protect\citeauthoryear{Shalit, Johansson, and
  Sontag}{2017}]{shalit2017estimating}
Shalit, U.; Johansson, F.~D.; and Sontag, D.
\newblock 2017.
\newblock Estimating individual treatment effect: generalization bounds and
  algorithms.
\newblock In {\em ICML}.

\bibitem[\protect\citeauthoryear{Tao \bgroup et al\mbox.\egroup
  }{2018}]{Tao2018}
Tao, C.; Chen, L.; Henao, R.; Feng, J.; and Duke, L.~C.
\newblock 2018.
\newblock {Chi-square Generative Adversarial Network}.

\bibitem[\protect\citeauthoryear{Thoemmes and Ong}{2016}]{Thoemmes}
Thoemmes, F., and Ong, A.~D.
\newblock 2016.
\newblock {A Primer on Inverse Probability of Treatment Weighting and Marginal
  Structural Models}.
\newblock {\em Emerging Adulthood} 4(1):40--59.

\bibitem[\protect\citeauthoryear{Thrusfield}{2017}]{Rosenbaum2002}
Thrusfield, M.
\newblock 2017.
\newblock {Observational studies}.
\newblock In {\em Veterinary Epidemiology: Fourth Edition}. Springer.
\newblock  319--338.

\bibitem[\protect\citeauthoryear{Yoon, Jordon, and van~der
  Schaar}{2018}]{yoonJS18}
Yoon, J.; Jordon, J.; and van~der Schaar, M.
\newblock 2018.
\newblock {GANITE:} estimation of individualized treatment effects using
  generative adversarial nets.
\newblock In {\em ICLR}.

\end{thebibliography}
\bibliographystyle{aaai.bst}

\end{document}